# Who to Trust, How and Why: Untangling AI Ethics Principles, Trustworthiness and Trust

**Andreas Duenser**, CSIRO, Australia, andreas.duenser@csiro.au
**David M. Douglas**, CSIRO, Australia, david.douglas@csiro.au

## Abstract

We present an overview of the literature on trust in AI and AI trustworthiness and argue for the need to distinguish these concepts more clearly and to gather more empirically evidence on what contributes to people's trusting behaviours. We discuss that trust in AI involves not only reliance on the system itself, but also trust in the developers of the AI system. AI ethics principles such as explainability and transparency are often assumed to promote user trust, but empirical evidence of how such features actually affect how users perceive the system's trustworthiness is not as abundance or not that clear. AI systems should be recognised as socio-technical systems, where the people involved in designing, developing, deploying, and using the system are as important as the system for determining whether it is trustworthy. Without recognising these nuances, 'trust in AI' and 'trustworthy AI' risk becoming nebulous terms for any desirable feature for AI systems.

## Introduction

Greater awareness of the potential harms of practical applications of artificial intelligence (AI) systems has inspired various approaches to preventing these harms. Various stakeholders, including governments, corporations, and non-government organisations, have developed principles for AI ethics.[1] A common theme among these sets of ethics principles is the need for AI systems to be *trustworthy*.

While often not clearly distinguished in the literature, trust in AI and trustworthy AI should be considered separately. While ideally trust in AI would be determined by the AI's trustworthiness, this correspondence between users' trust and AI trustworthiness is not automatic. Trustworthiness is a property of the AI system, while trust is an attitude that must be granted by users. Furthermore, users must properly calibrate their trust so that they do not distrust a trustworthy AI or trust an untrustworthy one. Trustworthiness is not the only factor that affects whether users decide to trust an AI. Focusing on the trustworthiness of AI systems without also considering the factors that influence how users develop trust in them will not achieve the goals of trustworthy AI. For this we also need to consider the wider AI system stakeholder ecosystem and the user's attitudes towards them as people may not trust or distrust the AI itself, but rather those *responsible* for it.

We begin this paper with an overview of the concepts of trust and trustworthiness, and how trust in AI differs from trust relationships between people. We then describe how trust is discussed in the context of AI ethics principles based on a short survey of AI papers that mention trust in the context of these principles. We highlight several issues, including an imprecise use of the terms trust and trustworthiness, a lack of clear empirical evidence with respect to if, how, and under what circumstances adhering to certain principles contributes to the formation of trust, and the need for a more nuanced consideration of the multidimensionality of AI ethics and potential trade-offs between principles. This leads to a discussion of trust in AI as socio-technical systems, and the need to consider the wider context of AI system development and use to better understand trust in AI.

## Trust

Trust itself has various forms. Interpersonal trust has been described as "if A [the trustor] believes that B [the trustee] will act in A's best interest, and accepts vulnerability to B's actions, then A trusts B".[2] Other forms of trust share significant characteristics with interpersonal trust.[3] For example, trust in organisations has been defined as "the willingness of a party [the trustor] to be vulnerable to the actions of another party [the trustee] based on the expectation that the other will perform a particular action important to the trustor, irrespective of the ability to monitor or control that other party".[3] Trust in automation has been defined as "the attitude that an agent will help achieve an individual's goals in a situation characterized by uncertainty and vulnerability".[4] Trust can also be seen as being contractual, i.e., the trustor believes that the trustee will act in accordance with and uphold a contract.[2]

These definitions share several common features with interpersonal trust. Trust occurs in a context of uncertainty: the trustee may not act in the way the trustor expects. Trust is an *attitude* that a trustor has towards a trustee, and it is limited to a particular context. Trust carries the risk that the trustee will not act as expected or as they promised: this is a betrayal of trust. The possibility of betrayal means that the trustor must determine whether trusting the trustee is an acceptable risk. This evaluation may be guided, in the case of interpersonal trust, by the trustor's perceptions of the trustee's ability, benevolence, and integrity: these characteristics represent the trustee's *trustworthiness*.[3] The trustor may be mistaken in their perceptions of the trustee's trustworthiness: a trustworthy person may be distrusted, and *vice versa*.

The three major Western theories of ethics (deontological ethics, consequentialism, and virtue ethics) all give justifications for the importance of maintaining the trust others place in us.[11] Deontological ethics, with its focus on acknowledging and fulfilling our duties to ourselves and to others, would regard being trustworthy to be important moral duty, and would condemn betraying another's trust unless another duty requires them to do so. Consequentialist ethics, which evaluates actions by whether they are likely to be beneficial or harmful to those they affect, will justify maintaining trust relationships if they are beneficial to both the trustor and trustee, and these benefits outweigh the potential harms to them both if the trustee betrays this trust. Virtue ethics focuses on developing and acting in accordance with intrinsically valuable character traits, such as courage, integrity, and trustworthiness itself. It also cautions against being too trusting of others (i.e., trusting those who are untrustworthy), and not trusting those who are worthy of it.

Technology, as a trustee, cannot betray the trustor in the same way as a human trustee can. The morality of trustors and trustees will influence how they trust and how they act when others trust them; moral considerations do not influence the actions of technology. In contrast to trusting a person (where the trustor's beliefs about the abilities, benevolence, and integrity will affect whether they trust a potential trustee), placing trust in technology depends on the trustor's beliefs about its technical characteristics.[5] This difference has led some to argue that technology (and other inanimate objects) cannot be trusted, but only relied upon: technology is *reliable* rather than trustworthy.[2] This objection may be answered by recognising the 'duality of trust' in technology, where humans 'trust' (or rely) on the technology itself and trust the technology supplier.[6] 'Trust' (or reliance) in technology may be guided by perceptions of its functionality (whether it is expected to be capable of performing the designated task), helpfulness (in the sense of being user-friendly and having help functions available), and reliability (whether it is expected to operate consistently and predictably).[5]

In automation technologies, it is possible to overtrust the technology if the user's perception of its capabilities exceeds what it can do.[4] Conversely, the technology may be distrusted if the user underestimates its capabilities. Overtrust may lead to misusing the technology (e.g., because of unwarranted trust people might use it even if it is not safe to do so), while distrust may lead to disuse.[4] In each case, the user's trust does not match the technology's trustworthiness. Calibrated, or warranted trust occurs when the user's trust in the technology accurately reflects its trustworthiness.[2,4]

## Trust in AI

AI systems that incorporate machine learning (ML) may identify patterns in the training data and use these patterns to derive algorithms for making decisions. As these algorithms are dependent on the system's training data rather than explicitly created by the developers, unanticipated behaviour may be a greater possibility than it may be for other computer systems. Users may also assign intention to AI systems to a greater degree than they would for other technologies, and this perception of intention may be used to justify referring to a willingness to depend on AI systems as 'human-AI trust' rather than reliance.[2] However, perceiving AI systems as having intentions risks perceiving them as moral agents, which implies that they are ethically responsible for their decisions and actions.[7] Attributing ethical responsibility to AI systems obscures the ethical responsibilities of AI developers, and potentially allows developers to deny responsibility and avoid accountability for the systems they create. AI systems may be causally responsible for decisions or actions, but the AI developers are ethically responsible for them.[7]

Similar to other conceptualisations and models of trust in AI (and/or automation), a recent literature review of user trust in AI identifies three major themes in the influences on user trust: socio-ethical considerations, technical and design features, and the user's own characteristics.[8] Socio-ethical considerations are factors in the social and ethical context where the system is used, such as user data protections, open communication with users, and clear accountability for potential user harms from the system.[8] Technical and design features include anthropomorphic design elements, explanations of how the system reached a decision, and the credibility of the system's decisions.[8] The user characteristics include inherent characteristics (such as personality traits), acquired characteristics (e.g., education and experience with AI), user attitudes (e.g., acceptance, expectations, and perceptions), and external variables (e.g., initial interactions, cognitive load, and the time spent with the system).[8]

An account of trustworthiness in explainable AI (XAI) models describes three factors that affect users' perceptions of trustworthiness in an AI system: performance, purpose, and process.[9] These factors are the user's understanding of the effectiveness of the AI system at performing its intended purpose, their understanding of the goal and motivation for developing it, and their understanding of how it works, respectively.[9] The transparency of an AI system is intended to provide users with the information needed for them to gain these understandings.[9]

## Different Types of Trust

We can differentiate various types of trust that may be relevant to trust in AI: dispositional, rational, affinitive, and procedural.[10] Each type may be fostered by social and ethical considerations, technical and design features of the AI system, and the user's characteristics.

Dispositional trust (or propensity to trust) is someone's tendency to trust or distrust others in particular contexts.[10] It is the baseline level of trust that exists before trustors develop rational or affinitive trust in trustees, and may be based on the trustor's personality, past experiences, and their understanding of cultural norms and context.[10] Dispositional trust in technology may be described as a general willingness to depend on technology (or not), and the willingness to trust (or distrust) a specific technology for a specific purpose.[5] The cultural group that a user belongs to may also affect their willingness to trust an AI.[11]

Rational (or cognitive) trust is the deliberative calculation of the beneficial outcomes of trusting another.[10] It requires the trustor to have sufficient information about the trustee to decide whether it will be beneficial to trust them. The trustor can give justifications for why they decided to trust the trustee, such as how they have performed when they have previously been trusted.[10,12] Rational trust in technology (such as AI) may be founded on the user's perceptions of the system's performance, purpose, and process.[9]

Affinitive (or emotional) trust is a judgement of whether to trust another.[10] Affinitive trust in technology may emerge from perceptions that it is user-friendly and reliable. User-friendly technology may foster a positive emotional response (through an attractive and comforting user interface), and perceptions of reliability may foster positive feelings of safety.

Procedural trust relies on procedures and other formal systems to limit the risk of trusting others.[10] A control system (such as an accepted set of procedures) is a means of establishing the trustworthiness of potential trustees: a trustee signals that they are trustworthy by announcing that they comply with a set of procedures that are relevant to potential trustors. Procedural trust may develop when both trustees and trustors accept the control system as legitimate.[10] However, the presence of a control system can also negatively affect the trustee's perceived trustworthiness, as their actions may be seen as a response to the control system rather than their inherent trustworthiness.[3] A control system that does not undermine the trustee's perceived trustworthiness is a positive control system.[10]

Distinguishing between these different types of trust helps to explain why we might use a technology while at the same time distrusting its creators. We may have affinitive distrust for a company that we perceive as having different values to our own, while also having dispositional or rational trust toward individual technologies it creates or procedural trust in the regulations it complies with. We might also be dependent on or have limited choice in not using certain technologies from developers we do not trust due to other factors (e.g., we must use such technologies in a work setting).[8]

This illustrates that trust is multidimensional and that trust in AI may be considered not only from the user-system perspective but should also take the broader stakeholder environment into account. Having affinitive trust in a technology creator may impact our trust towards their technologies. Similarly, procedural trust in the regulations and procedures (such as AI ethics principles) that a developer adheres to also may impact a person's trust towards their technologies.

## Trust and AI Ethics Principles

The potential negative impacts of AI systems on users has led, as many emphasise, to user distrust in these systems, and highlighted the importance of AI ethics.[8,12] One approach to AI ethics that draws on medical ethics is principlism, where a set of ethical principles provide a common language and broad guidance for discussing ethical issues.[13] AI ethics principles and guidelines have been developed to provide normative guidance on how to apply ethics considerations to the design of AI-based systems.[1,12] These have been discussed as one of the socio-ethical factors that may influence user trust in AI.[8]

A rough consensus is emerging among the many sets of AI ethics principles published globally.[1] An analysis of 36 documents describing AI ethics principles identified eight major themes: privacy, accountability, safety and security, transparency and explainability, fairness and non-discrimination, human control of technology, professional responsibility, and promoting human values.[1] These principles describe factors that can be relevant to a system's trustworthiness that, in turn, may foster particular forms of trust. Privacy, fairness and non-discrimination, human control of technology, and promoting human values may foster affinitive trust in the AI developer by indicating that they respect and share the values of users, and that these values shape the design and operation of the AI. Accountability, safety and security, and transparency and explainability could help developing rational trust in AI.[9] Each of these principles indicate that there are procedures that guide the design and operation of the AI and serve as control systems that may encourage procedural trust.

While one of the purposes of following AI ethics principles in the development of AI systems may be to foster procedural trust in AI, the main goals of these principles are to identify and (if adhered to) minimise an AI's capacity to cause harm to people, and to provide people with information they need to make decisions about exposing themselves to the risk of that harm.[14] Reinhardt argues that the current discussion about AI ethics 'overloads' the notions of trust and trustworthiness, and turns trust into a buzzword and umbrella term for an inconclusive list of things deemed 'good'.[15] This overloading, a lack of consideration of potential contradictions between ethics principles, and little consideration of the targeted nature of trust highlights another element of potential confusion: must AI systems adhere to all AI ethics guidelines to be trusted? Must only some principles be adhered for trust to emerge? Which principles, or combinations of principles, might be more important or necessary than others for building trust?

To get a clearer idea how trust is discussed or evaluated in the context of AI ethics principles, we reviewed a subset of papers that were originally collected as part of a broader systematic literature review on design patterns for responsible AI conducted by our wider project team.[16] After performing full-text searches for "trust" on these papers, we performed an initial scan and discarded papers that may have mentioned trust but did not provide any substantial detail on the topic. This resulted in 172 papers which we reviewed in more depth and categorized according to the principle(s) listed in the Australian AI Ethics Principles that were most relevant to how trust and/or trustworthiness were discussed in each paper (see Table 1).[17]

We found that trust and trustworthiness are often not clearly separated and repeatedly used interchangeably. This conflation can be problematic as it is important to keep "trust" (an attitude of the trustor) distinct from being "trustworthy" (a property of the trustee).[2,15] While AI ethics guidelines may describe elements of a system's trustworthiness, it is not a given that users will actually trust a trustworthy system. Trustworthiness is not a necessary antecedent or prerequisite for trust and a trustworthy system does not necessarily lead to trust in users.[2] This distinction between trust and trustworthiness means that focusing on the trustworthiness of AI systems is insufficient: trust has to be *granted* by the trustor.[15] Elements that contribute to a system's trustworthiness (such as signalling that a system conforms to certain standards or principles) may be irrelevant for a trustor's perception or judgement of such a system and fail to solicit the appropriate trust response.

**Table 1.** Papers discussing "Trust" in the context of the eight Australian AI Ethics Principles (N = 172). Note: papers may cover more than one AI ethics principle.

| AI Ethics Principle | Papers Mentioning Trust (N=172) |
|---|---|
| Transparency and Explainability | 108 |
| Accountability | 23 |
| Reliability and Safety | 23 |
| Privacy Protection and Security | 21 |
| Fairness | 20 |
| Human, Social and Environmental Wellbeing | 11 |
| Human-Centred Values | 9 |
| Contestability | 8 |

Trust is targeted and contextual: we trust a system to do or achieve something specific or to behave in a certain way. For example, we may expect (and trust) a medical decision support system to provide accurate medical advice. However, we may not expect (and trust) the same system to provide accurate legal advice. From the perspective of trust as being a contract between the trustor and trustee, and depending on a person's beliefs, priorities or other factors, certain AI ethics principles may or may not be part of their "trust-contract" in certain contexts or more generally. Hence, specific principles or combination of principles may or may not affect a system's trustworthiness or capacity to create trust for everybody in the same way (or at all) in each context.

Table 1 shows that within our sample of reviewed papers, trust was most often mentioned in the context of transparency and explainability. The general assumption in many of these papers was that explainability and/or transparency will "build trust". However, the empirical evidence, where available, suggests that this assumption cannot always be taken for granted. While explanations have been shown to increase users' willingness to trust AI, in other cases transparency may have negative effects on trust in an AI.[9,18]

In line with other work, we found that the connection between trust and in AI ethics principles is often assumed rather than described in detail or underpinned by empirical evidence[8]. Within the sample of papers we reviewed, empirical results (where they existed) were unclear about when explainability and transparency contribute to trust, in which contexts they may do so, or what types of explainability are useful for specific stakeholders to trust an AI, and whether these are the only or most important factors for

(warranted) trust to form.

Many of the papers that we reviewed that discussed explainability took a rather technical perspective with regards to how it should be implemented (e.g., determining which features contribute to a model's output), but mostly did not examine whether users could fully understand such explanations and develop a better understanding of the AI and/or its outputs. The usefulness of an explanation depends on the user's background knowledge, and for an explanation to be effective in developing trust, it must be appropriate for its intended users. The actual trustworthiness of explanations themselves (i.e., is the explanation accurate or reliable?) was also rarely discussed. Explanations can be misleading, either unintentionally or even intentionally. There is a risk, especially if explanations are given to foster trust, that a user's beliefs could be manipulated to align with a specific explanation of a phenomenon, and thus users could be nudged to form a preferred belief or take a preferred action.[19]

It has been argued that transparency might be more useful to manage a lack of trust rather than for creating trust in a system.[15] When we ask for more transparency, we ask for more control and reduce the actual need for trust as it reduces the perceived risk and uncertainty about the AI's actions.[15] While explainability still requires some level of trust in the AI developer that the explanations accurately represent how the AI operates, transparency reveals the actual operation of the system rather than a representation of it. Transparency may remove the expected cost of giving up control over an action or decision to the AI.[20] It also reduces the duality of trust in AI into reliance in the system as transparency reduces the need to trust the developer.

Trust and trustworthiness were less frequently discussed in the context of other AI ethics principles as the numbers in table 1 illustrate. As with transparency and explainability, we found little empirical evidence within the papers we reviewed regarding if and how exactly responding to the concerns represented by these principles may impact trust in users.

Although some papers considered subjects relevant to multiple ethics principles in the context of trust, most focused on specific principles in isolation. Considering that AI ethics is multidimensional, a more nuanced analysis of trustworthiness and trust is needed, and we need to consider potential trade-offs between certain principles. A system might be trustworthy with respect to explainability and transparency, but not necessarily with respect to human-centred values, contestability, or privacy.[15] Similarly, an AI-supported recruitment system might be fair with respect to certain demographic features but biased with respect to others, and there might be issues with accountability and transparency.

What this means for individuals deciding whether to trust these systems is not entirely clear, and clear evidence to answer such questions is lacking. Some may put more weight on specific aspects or types of fairness than other types. Others might have varying levels of concern about transparency, accountability and/or other ethics principles when evaluating the AI's trustworthiness. All this may affect whether users grant trust. It is important to discuss issues of individual weighing of principles or of conflicts between different principles in more detail. We should not turn AI ethics into what Reinhardt calls an "intellectual land of plenty" by using these concepts as "umbrella terms for everything that would be nice to have regarding AI systems, both from a technical as well as an ethical perspective".[15]

## Trust in AI as Socio-Technical Systems

When exploring trust in AI, we must consider the use context of these systems, who is using them, the purposes for which they are developed and used as well as the wider stakeholder ecosystem.

AI systems are socio-technical systems, where the AI is itself part of a larger system of people, laws, technologies, institutions, and social norms.[15] AI ethics principles themselves are also socio-technical systems and may serve as controls intended to establish procedural trust in the AI systems created by developers who follow the principles. The effectiveness of ethical AI principles will follow from their perceived legitimacy from both AI developers and users. One problem for AI ethical principles as a control system is the limited legal and professional accountability mechanisms for AI developers.[13] Without effective governance bodies and penalties for developers who fail to adhere to AI ethics principles, their effectiveness as a means of establishing procedural trust may be limited.[13]

An important point that is often overlooked in the discussion of ethical AI is that responsibility or ethics pertain to the socio-technical context of how technology is developed and used, rather than to the specifics of the tool itself. Kroll highlights this point by discussing the inscrutability of algorithms as an example.[14] ML algorithms that are inscrutable or 'black boxes' that cannot be easily understood (if at all) are mostly presented as an issue with the technology itself, rather than the result of decisions made by the developers who created them. However, such algorithms and the systems built with them are products of a socio-technical context. Inscrutability is therefore not merely a result of technical complexity but of choices made by the system's designers, developers, operators and controllers.[14] Technology systems may be explained and understood with respect to their designs and operational goals as well as their inputs, outputs, and outcomes. Whoever controls the deployment of an AI system may accept that inscrutable or opaque decisions are satisfactory for using that system for its intended purpose.[14] In such cases, the developer has decided that users would be willing to accept decisions made by an opaque AI system, or the user decided that the system is trustworthy or reliable enough to use despite its inscrutability. However, this view may not be shared by those affected by the decisions made using the AI system.

Those affected by the system's use may also differ in their understanding of the purpose of the AI system. This difference may lead to a system being distrusted by those affected by its use, despite its users considering it trustworthy. For example, a doctor may consider an AI-based decision support tool assisting them in patient diagnosis to be trustworthy, but the patient may not. Those affected by the system's use must depend on the trustworthiness of the system user, the trustworthiness of the AI itself for its intended purpose, and the trustworthiness of the AI system provider.

Trust transfer also highlights the importance of considering the stakeholder ecosystem.[6] It describes a situation where trust in an entity may transfer to a novel target (e.g., trust is transferred from known technologies and/or providers to others). Trust in a new target may be influenced by the perceived relationship between the trust source and the target. If a person trusts A and perceives that A has a relationship with B, the person may also trust B. The 'duality of trust' (reliance in the technology itself and trust in the technology provider) is particularly relevant when considering trust in AI. People often have little understanding of specific AI technologies or may not even be aware that they are using such systems, yet they are likely more familiar with the corporations providing the technologies. Trust transfer explains how technology users transfer their trust not only from known technologies to new technologies, but also from trust (or distrust) in technology providers to the technologies they develop.

Dispositional trust and affinitive trust in the technology provider (due to brand recognition, reputation, or positive experiences with other products) may transfer to the AI integrated into their products. In such cases, the user determines that the AI can be trusted despite not having rational trust in it, as they may lack the knowledge and understanding to make a well-informed decision about its trustworthiness.[7] Trust in an AI that is based on the user's trust in the system's developer is interpersonal or organisational trust rather than human-AI trust, as the user trusts the developer rather than the AI itself.[2] While human-AI trust has been described as the user's perception that an AI is trustworthy with regard to fulfilling an explicit contract and accepting the vulnerability of relying on it to do so, this definition obscures the importance of the user's perception of the trustworthiness of the AI developer. While an AI may be trustworthy regarding completing an explicit contract, part of this may derive from the perceived trustworthiness of the AI developer.

Many models of trust in automation and trust in AI (implicitly) consider a person (trustor) directly interacting with an AI system (trustee). In cases where such interaction is mediated (e.g., through an ecommerce or social media platform, for example), the trust relationship also involves these mediators and is impacted by procedural trust. Trust transfer also may come into play here, where trust or distrust in the mediator (e.g., ecommerce platform provider, social media company) can transfer to the underlying technology (e.g., a recommendation algorithm). This again points to the need for considering the wider stakeholder ecosystem and socio-technical context of technology use.

## Conclusion

To allow us to move from principles to practice, we need to further untangle and develop our understanding of how AI ethics principles, that might inform the development of AI systems, relate to individual perceptions of trustworthiness, and how such trustworthiness matters (or if at all, and under what circumstances) to people and impacts trust.

AI ethics incorporates several dimensions which are underscored by the various principles that have been developed. Trust is targeted and context dependent. Depending on this context, multiple of these dimensions may contribute to a person's trust evaluations and trusting behaviours when interacting with an AI. We need more empirical evidence to better understand how and if certain AI ethics principles, combinations of principles, and potential trade-offs or conflicts between principles impact people's trust. For some, certain principles might be more important than others depending on their individual circumstances or experiences. Thus, along with others, we argue that these issues require more nuanced discussion and empirical evaluation so that trust and trustworthiness do not degrade into vague catch-all terms for the 'good' in AI systems.

We need to consider AI as socio-technical systems, where the dynamics between various stakeholders may impact on perceptions and evaluations of trustworthiness, and how (and under which circumstances) such factors may contribute to creating or eroding trust as well as affecting the acceptance and adoption of AI. We need to pay attention to how the relationships and dynamics between various stakeholders, their trustworthiness and whether they adhere (or are perceived to adhere) to AI ethics principles, can affect trust and trustworthiness perceptions of AI systems, developers, providers and users, and how this affects the acceptance or adoption of AI technologies.

The importance of the wider stakeholder ecosystem highlights the significance of AI ethics principles that guide the relationships between stakeholders. Principles such as accountability and contestability (from the Australian AI Ethics Principles), and the theme of professional responsibility from the review of global AI ethics principles are as important for fostering trustworthiness as other principles such as safety, transparency and explainability.[1,17] Trustworthiness of AI systems should reflect aspects of the AI itself and the trustworthiness of the stakeholders associated with the system, such as the developers of the system itself and those who incorporate AI into their professional practice.

## Acknowledgements

We thank Qinghua Lu and Conrad Sanderson for their work in the structured literature review that we drew upon in section 2, and Melanie McGrath for her helpful comments. This research was funded by Data61 and CSIRO's Responsible Innovation Future Science Platform.